%% file: main.tex
\documentclass[conference]{IEEEtran}
\IEEEoverridecommandlockouts

\usepackage{cite}
\usepackage{amsmath,amssymb,amsfonts}
\usepackage{algorithm}
\usepackage{algpseudocode}
\usepackage{graphicx}
\usepackage{textcomp}
\usepackage{xcolor}
\usepackage{multirow}
\usepackage{cite}
\usepackage{hyperref}
\usepackage{cleveref}

\DeclareMathOperator*{\argmin}{arg\,min}

\def\BibTeX{{\rm B\kern-.05em{\sc i\kern-.025em b}\kern-.08em
    T\kern-.1667em\lower.7ex\hbox{E}\kern-.125emX}}
\begin{document}

\title{PG-DPIR: An efficient plug-and-play method for high-count Poisson-Gaussian inverse problems\\
\thanks{This work was partly supported by CNES under project name DEEPREG, and ANITI under grant agreement ANR-19-PI3A-0004.}
}
\author{\IEEEauthorblockN{Maud Biquard}
\IEEEauthorblockA{
\textit{ISAE-Supaero/CNES}\\
Toulouse, France \\
maud.biquard@isae-supaero.fr}
\and
\IEEEauthorblockN{Marie Chabert}
\IEEEauthorblockA{
\textit{IRIT / Toulouse-INP}\\
Toulouse, France \\
marie.chabert@irit.fr}
\and
\IEEEauthorblockN{Florence Genin, Christophe Latry}
\IEEEauthorblockA{
\textit{CNES}\\
Toulouse, France \\
firstname.lastname@cnes.fr}
\and
\IEEEauthorblockN{Thomas Oberlin}
\IEEEauthorblockA{
\textit{ISAE-Supaero}\\
Toulouse, France \\
thomas.oberlin@isae-supaero.fr}
}

\maketitle

\input{sec/0_intro}
\input{sec/1_contrib}

\input{sec/2_experiments}
\input{sec/3_conclusion}




\bibliographystyle{ieeetr}
\bibliography{IEEEabrv,biblio}

\end{document}

%% file: sec/0_intro.tex
\begin{abstract}
Poisson-Gaussian noise describes the noise of various imaging systems thus the need of efficient algorithms for Poisson-Gaussian image restoration. 
Deep learning methods offer state-of-the-art performance but often require sensor-specific training when used in a supervised setting. A promising alternative is given by plug-and-play (PnP) methods, which consist in learning only a regularization through a denoiser, allowing to restore images from several sources with the same network. This paper introduces PG-DPIR, an efficient PnP method for high-count Poisson-Gaussian inverse problems, adapted from DPIR \cite{Zhang2021}. While DPIR is designed for white Gaussian noise, a naive adaptation to Poisson-Gaussian noise leads to prohibitively slow algorithms due to the absence of a closed-form proximal operator. To address this, we adapt DPIR for the specificities of Poisson-Gaussian noise and propose in particular an efficient initialization of the gradient descent required for the proximal step that accelerates convergence by several orders of magnitude. Experiments are conducted on satellite image restoration and super-resolution problems. High-resolution realistic Pléiades images are simulated for the experiments, which demonstrate that PG-DPIR achieves state-of-the-art performance with improved efficiency, which seems promising for on-ground satellite processing chains.
\end{abstract}


\section{Introduction}

Poisson-Gaussian noise is the combination of Poisson noise, related to the quantum nature of light, and of Gaussian noise, modeling the electronic noise of the imaging system. Hence, this noise model exists in various applications, such as remote sensing, astronomy, or biology. The forward model of such a system can be described as:
\begin{equation}
    y = p(Ax) + w_0 \label{eq:pgnoise} 
\end{equation}
with $x$ the target, $A$ a degradation operator (e.g. a convolution), $p(Ax) \sim \mathcal{P}(Ax)$ some Poisson noise, and $w_0 \sim\mathcal{N}(0,\sigma_0^2I)$ some Gaussian noise.
Then, the imaging problem consists in recovering $x$, typically by solving
\begin{equation}
    \arg\min_x F(x,y) + \lambda R(x) \label{eq:optim}
\end{equation}
with $F(x,y)$ measuring the fidelity between $Ax$ and $y$, and $R$ being a regularization or a prior on $x$, promoting solutions most compatible with $R$.

Except for low-count imaging, it is realistic to perform a Gaussian approximation of the noise model in \cref{eq:pgnoise} \cite{Foi2008, Li2015}, leading to the following forward model:
\begin{equation}
    y = Ax + w, \quad w \sim \mathcal{N}(0,\sigma_w^2(x)) \label{eq:pgapprox}
\end{equation}
with $\sigma_w^2(x)=\sigma_0^2 + K (Ax)$,  $\sigma_0$ and $K$ being the noise parameters. 
This approximation is very accurate as long as the number of photons exceeds 20. This condition is satisfied in various applications, for instance for remote sensing problems \cite{Latry2012}.
To solve such inverse problems, classical approaches often rely on the Anscombe transform \cite{Makitalo2012} which empirically stabilizes the noise variance. 

However, deep learning methods have significantly improved image restoration performance.  
The majority of these methods \cite{Ledig2017,Zhang2018a} consists in directly mapping the degraded image to the restored image using supervised learning, for a specific inverse problem. This requires one neural network for each sensor, making this unpractical for real-world processing pipelines with several sources. Unlike them, plug-and-play (PnP) methods consist in solving \cref{eq:optim} with proximal splitting algorithms and replacing the prior step by a denoising step \cite{Zhang2021,Venkatakrishnan2013,Hurault2022}. The denoiser regularizes the problem independently from the forward model, enabling to restore images from several sources within the same network. This makes them interesting for replacing traditional methods in the processing chains, however, these methods often require lots of iterations to converge, making them less competitive than supervised methods in terms of computation time. Interestingly, DPIR \cite{Zhang2021} is a PnP method requiring very few iterations as it leverages a denoiser trained at several noise levels to speed up the convergence. Yet, DPIR is designed for white Gaussian noise for which the prox can be computed in closed form. Thus, it can become very slow when the prox is not available in a closed form, as it has to be computed within gradient descent. This is the case for the considered inverse problems.

In this paper, we introduce PG-DPIR, an adaptation of DPIR for Poisson-Gaussian image restoration. 
In this method, besides adapting DPIR hyperparameters for Poisson-Gaussian noise, we propose an efficient procedure to compute the data-fidelity prox by initializing the required gradient descent with an approximated proximal operator that can be computed in closed form. Afterward, only few gradient steps are required, leading to a very fast algorithm.
We apply PG-DPIR to satellite image restoration. 
In particular, we compare PG-DPIR to algorithms that are currently used in real-world pipelines, as well as to various PnP methods. These experiments are conducted on realistic images simulated from airplane images to imitate the very high resolution Pléiades satellites. These experiments show that PG-DPIR outperforms the other PnP methods in terms of metrics and computation speed, showing the potential of integrating PG-DPIR in satellite image processing pipelines.  

\section{Related works}

Direct inversion methods for image restoration directly map the degraded to the restored image in a supervised manner. They rely on diverse neural architecture, which can be in particular convolution-based \cite{Ledig2017,Wang2018}, transformer-based \cite{Wang2022}, or diffusion-based \cite{Luo2023,Lugmayr2022}. 
Unlike them, the methods that only learn the regularization enable to solve several inverse problems with the same network. This regularization can be learned explicitly within generative models \cite{Bora2017, Biquard2023a}, or implicitly with a denoiser \cite{Venkatakrishnan2013,Zhang2021} in PnP methods or with a diffusion model \cite{Zhu2023,Kawar2022} in diffusion-based methods. Concerning PnP methods, they solve \cref{eq:optim} using diverse splitting algorithms, such as Alternating Direction Method of Mutlipliers (ADMM) \cite{Venkatakrishnan2013}, Half Quadratic Splitting (HQS) \cite{Zhang2021} or Proximal Gradient Descent \cite{Hurault2023a}. In particular, DPIR \cite{Zhang2021} uses the HQS framework, solving \cref{eq:optim} with very few iterations.

Although Poisson noise is well studied in the literature \cite{Rond2016,Syed2023}, Poisson-Gaussian noise is less frequently addressed. Some approaches consider the exact Poisson-Gaussian noise model. For instance, \cite{Chouzenoux2015} formulates the exact Poisson-Gaussian likelihood and considers a primal-dual algorithm while \cite{Lanza2014} solves a minimax problem using a total variation regularization. This often leads in practice to heavier and more complex methods than considering the approximated Gaussian model of \cref{eq:pgapprox} \cite{Benvenuto2008}. \cite{Li2015} considers the approximate noise model and proposes a variational restoration algorithm using framelet transform, and \cite{Benvenuto2008} performs a Expectation-Maximization method with this same noise approximation. \cite{Latry2012} proposes to restore satellite images using the Anscombe variance stabilization \cite{Makitalo2012} and then applies classical white Gaussian denoising and deconvolution techniques. Concerning deep learning methods, \cite{Zhang2019a} has trained a Poisson-Gaussian denoiser for fluorescence microscopy. Interestingly, \cite{Hurault2023} proposes a convergent PnP framework for Bregman noise models applied to Poisson noise that could potentially work for Poisson-Gaussian noise.

%% file: sec/1_contrib.tex
\section{Method}

\subsection{Background on DPIR}

DPIR \cite{Zhang2021} is a PnP algorithm relying on Half Quadratic Splitting (HQS). Its particularly low number of iterations as well as its convenient parameter tuning have made DPIR a reference PnP method for solving inverse problems with white Gaussian noise. 

\subsubsection{Optimization process}
Considering a forward model $y = Ax + w$ with $A$ the degradation operator and a white Gaussian noise $w \sim \mathcal{N}(0, \sigma^2)$, DPIR seeks to minimize $\frac{1}{2\sigma^2}||y - Ax||^2 + \lambda R(x)$ with respect to $x$ where $R$ is a regularization term.
HQS algorithm consists of the following splitting:
\begin{equation}
    \min_{x,u} \frac{1}{2\sigma^2}||y - Ax||^2 + \lambda R(u) + \frac{\mu}{2}||u-x||^2, \label{eq:hqs}
\end{equation}
with $u$ an auxiliary variable that is jointly optimized with $x$. This leads to an alternate optimization scheme
\begin{align}
    x_k &= \argmin_x \frac{1}{2\sigma^2}||y - Ax||^2 + \frac{\mu}{2} ||x - u_{k-1}||^2 \label{eq:prox_datafit}, \\
    u_k &= \argmin_u \frac{1}{2 \sqrt{\lambda / \mu}^2}||u - x_k||^2 + R(u)  \label{eq:prox_prior}.
\end{align}
\Cref{eq:prox_datafit} is a proximal step on the data fidelity which can be computed in closed form for various inverse problems. \Cref{eq:prox_prior} amounts to a denoising step with noise level $\sigma_d = \sqrt{\lambda/\mu}$. 

\subsubsection{Noise schedule}
\cite{Zhang2021} proposes to use decreasing values of $\sigma_d$, evenly spaced in log scale from $\sigma_1$ to $\sigma_2 < \sigma_1$ over the iterations. $\sigma_2=\sigma$ corresponds to the noise level in the measurement, while $\sigma_1 > \sigma$ is a hyperparameter that is typically set to $50/255$ for 8 bits images. With $\lambda$ a fixed regularization parameter, this leads to a larger $\mu = \frac{\lambda}{\sigma_d^2}$ at each iteration so that $u_k$ and $x_k$ become closer over the iterations. 
Note that the number of iterations is fixed \textit{a priori} because of the noise schedule.
This results in a very fast convergence, generally within 8 iterations.
In \cite{Zhang2021}, the successive denoising steps are performed with DRUnet denoiser as it can manage different noise levels. 

\subsection{Proposed Poisson-Gaussian DPIR}

DPIR algorithm has been specifically designed for inverse problems with white Gaussian noise of fixed variance. 
Hence, it has to be adapted to Poisson-Gaussian inverse problems. We consider the approximate forward model of \cref{eq:pgapprox}, for which the Poisson-Gaussian noise is approximated as a Gaussian but with a spatially varying variance $\sigma_w^2(x) = \sigma_0^2 + K (Ax)$. 
This forward model substantially complicates the problem resolution. 
In particular, the proximal step of \cref{eq:prox_datafit} can no longer be computed in closed form and has to be obtained through gradient descent. This significantly slows down the algorithm convergence for large images. To overcome this limitation, we propose in this section a smart initialization of this gradient descent that considerably speeds up its convergence.

First, for the considered inverse problem, the quadratic data fidelity term in equations \eqref{eq:hqs} and (\ref{eq:prox_datafit}) should be replaced by the real negative log-likelihood:
\begin{align}
    F(x,y) \propto \frac{1}{2}(y - Ax)^T &\Sigma_w(x)^{-1} (y - Ax) + \frac{1}{2} \log |\Sigma_w(x)|
\end{align}
up to a constant, with $\Sigma_w(x) = \operatorname{diag} (\sigma_w^2(x))$. Thus, we replace \cref{eq:prox_datafit} with:
\begin{align}
    x_k &= \arg \min_x F(x,y) + \frac{\mu}{2} ||x - u_{k-1}||^2. \label{eq:prox_datafit_adapted}
\end{align}
We adapt the noise schedule by considering $\sigma_2 = \sigma_0$, that is the minimum level of noise in the image. 
\begin{algorithm}
\small{
\caption{PG-DPIR}\label{alg:satdpir}
\begin{algorithmic}
    \Require $y$ measure, $u_0$ initial image, $D(x;\sigma_d)$ denoiser, $n$ iterations,  $(\sigma_d^{(1)},...,\sigma_d^{(n)}=\sigma_0)$ denoising schedule, $\lambda$ parameter, $\eta$ inner gradient descent stepsize
    \For{$i=1,...,n$}
      \State $\bar \sigma \gets \sqrt{a + b(h_0*\bar u_{k-1})}$ 
      \Comment{Mean noise level of $u_{k-1}$}
      \State $x_k^{(0)} \gets \arg \min_x \frac{1}{2 \bar \sigma^2}||y - h_0 * x||^2 + \frac{\lambda}{2 (\sigma_d^{(i)})^2} ||x - u_{k-1}||^2$
      \If{$i\leq\frac{n}{2}$}
        \State $x_k \gets x_k^{(0)}$ 
      \Else \Comment{Gradient descent}
        \For {$j=1,...,5$}
            \State $x_k^{(j)} \gets x_k^{(j-1)} - \eta \nabla_x [F(x,y) + \frac{\lambda}{2 (\sigma_d^{(i)})^2} ||x - u_{k-1}||^2]$ 
        \EndFor
        \State $x_k \gets x_k^{(5)}$
      \EndIf
      \State $u_k \gets D(x_k, \sigma_d^{(k)})$ \Comment{Denoising step}
    \EndFor
\end{algorithmic}}
\end{algorithm}
Then, \cref{eq:prox_datafit_adapted} can no longer be expressed in closed form, and should be computed using gradient descent. We propose to initialize the gradient descent at iteration $k$ with 
\begin{equation}
    x_k^{(0)} = \arg \min_x \frac{1}{2 \bar \sigma^2}||y - Ax||^2 + \frac{\mu}{2} ||x - u_{k-1}||^2 \label{eq:prox_datafit_init}
\end{equation}
where $\bar \sigma^2 = \sigma_0^2 + K(A\bar u_{k-1})$ and $\bar u_{k-1}$ is the mean value of $u_{k-1}$. \Cref{eq:prox_datafit_init}, which corresponds to the proximal operator considering a fixed noise level $\bar \sigma^2$, can be computed in closed form. Then, we consider two phases during DPIR iterations. During the first half of the iterations, where each iteration corresponds to significant changes in $x_k$, we do not perform gradient descent on \cref{eq:prox_datafit_init} but only the proximal step \cref{eq:prox_datafit_init}, that is $x_{k-1} = x_{k-1}^{(0)}$. Indeed, we consider $x_k^{(0)}$ to be a sufficiently good approximation to $x_k$ in this case. During the second half of the iterations, we perform gradient descent on \cref{eq:prox_datafit_adapted} starting from $x_k^{(0)}$ during only 5 iterations. The whole PG-DPIR algorithm is provided in \cref{alg:satdpir}. We show in the experiments section that this optimization strategy produces very similar results as the original process while being much faster.

%% file: sec/2_experiments.tex
\section{Experiments}

We conduct experiments on optical satellite image restoration problems on panchromatic images. For this application, we consider the following forward model:
\begin{align}
    y = \mathcal{D}(h*x) + w \label{eq:ip_sat},
\end{align}
where $x$ represents the observed landscape at the target resolution, $y$ the acquired image, $h$ the blur kernel sampled at the target resolution, modeling the combined effects of the atmosphere, of the movement during integration time, and of the instrument. $\mathcal{D}$ denotes a downsampling operator. In the experiments, we consider a restoration problem, for which $\mathcal{D}=Id$, and a joint restoration and super-resolution problem, for which $\mathcal{D}$ downsamples the image by a factor of 2.
Vector $w$ represents a white Poisson-Gaussian noise which can, for the considered problems, be approximated by a Gaussian noise of variance $\sigma_w^2(x)=\sigma_0^2 + K \mathcal{D}_0(h*x)$, where $\sigma_0$ and $K$ are noise parameters specific to a given optical system.

\subsection{Simulation of realistic satellite images}

We train and test the network on realistic simulated images, imitating images from Pléiades satellites. For the training, target images only are required, while for the image restoration, pairs of (target image, degraded image) are required. These data are simulated from airplanes images at extremely high resolution, allowing us to consider them as perfect when downsampled at the target resolution. We use two databases: PCRS, provided by IGN \cite{IGNPCRS}, and Pélican, provided by CNES. PCRS are 12 bits images, with a resolution of 5cm covering $537.18 km^2$. Pélican are 12 bits images with a resolution of 10cm covering $18.45km^2$.

To simulate the target images, the images are downsampled at the target resolution using a bicubic filter. The degraded images are simulated using a realistic simulation chain from CNES to imitate images from Pléiades satellites, which have a resolution of 50cm and for which the degradation model is well controlled by CNES. The test set is constituted of 30 images at a resolution of 50cm of size $820\times820$, coming from the Pélican dataset. The other images are used for training.

\subsection{Experiment setup}

\subsubsection{Considered inverse problems}
We consider two problems: Pléiades image restoration without, denoted as IR, and with super-resolution by 2, denoted as SISR.
\subsubsection{PG-DPIR details}
We use DRUNet denoiser \cite{Zhang2021}. It is a state-of-the-art deep convolutional denoiser designed for image restoration. We employ a pretrained model from \cite{Zhang2021} and finetune it on target images during 500k iterations. For IR, 8 PG-DPIR iterations are performed, and 20 iterations for SISR. We reduce the starting denoising level $\sigma_d^{(1)}$ to 20/255 (that is 320/4095 for 12-bits images) as the noise levels considered in these restoration problems are typically lower than in the original DPIR paper \cite{Zhang2021}.
\subsubsection{Baselines}
 We compare PG-DPIR to a classical image restoration algorithm currently used in Pléiades ground segment, that we denote Bay+IF. It includes NL-Bayes \cite{Lebrun2013} for denoising then inverse filtering. For SISR, the Bay+IF images are upsampled using a bicubic filter. We also use DPIR as a baseline, for which we consider an approximated white Gaussian noise model with the deviation equal to the Poisson-Gaussian deviation at the mean luminance. At last, we compare to proximal gradient descent (PGD), another PnP method. All the PnP methods are performed with the same denoiser. The hyperparameters of PG-DPIR and the baselines are tuned on a validation set of 4 images.
\subsubsection{Metrics}
We use three metrics: the Peak Signal-to-Noise Ratio (PSNR), the Structural SIMilarity (SSIM) \cite{Wang2004}, and the Learned Perceptual Image Patch Similarity (LPIPS) \cite{Zhang2018}. The PSNR represents the accuracy between the restored and target images, while the SSIM and LPIPS are respectively classical and deep learning perceptual metrics.

\subsection{Results}

\begin{table}[h]
\caption{Results for the image restoration (IR) and super-resolution (SISR) problems. Time denotes the time required to restore one image with a Nvidia Quadro RTX8000 GPU.}
\setlength{\tabcolsep}{5pt}
\centering
\begin{tabular}{clcccc|c}
\hline
\multicolumn{1}{l}{} &  & $y$ & \textbf{Bay+IF} & \textbf{DPIR} & \textbf{PGD} & \textbf{PG-DPIR} \\ \hline
\multirow{4}{*}{\textbf{IR}} & PSNR $\uparrow$ & 33.55 & 40.56 & 48.52 & 45.90 & \textbf{48.66} \\
 & SSIM $\uparrow$ & 0.9289 & 0.9859 & 0.9950 & 0.9925 & \textbf{0.9952} \\
 & LPIPS $\downarrow$ & 0.1437 & 0.0369 & 0.0163 & 0.0428 & \textbf{0.0138} \\
 & Time & x & x & \textbf{2.2s} & 21.1s & 3.7s \\ \hline
\multirow{4}{*}{\textbf{SISR}} & PSNR $\uparrow$ & x & 33.22 & 37.04 & 34.55 & \textbf{37.18} \\
 & SSIM $\uparrow$ & x & 0.9088 & 0.9505 & 0.9319 & \textbf{0.9513} \\
 & LPIPS $\downarrow$ & x & 0.2463 & 0.1676 & 0.2058 & \textbf{0.1658} \\
 & Time & x & x & \textbf{60.5s} & 204.2sec & 68s \\ \hline
\end{tabular}
\label{tab:metrics}
\end{table}

\begin{figure}[h]
    \centering
    \includegraphics[width=\linewidth]{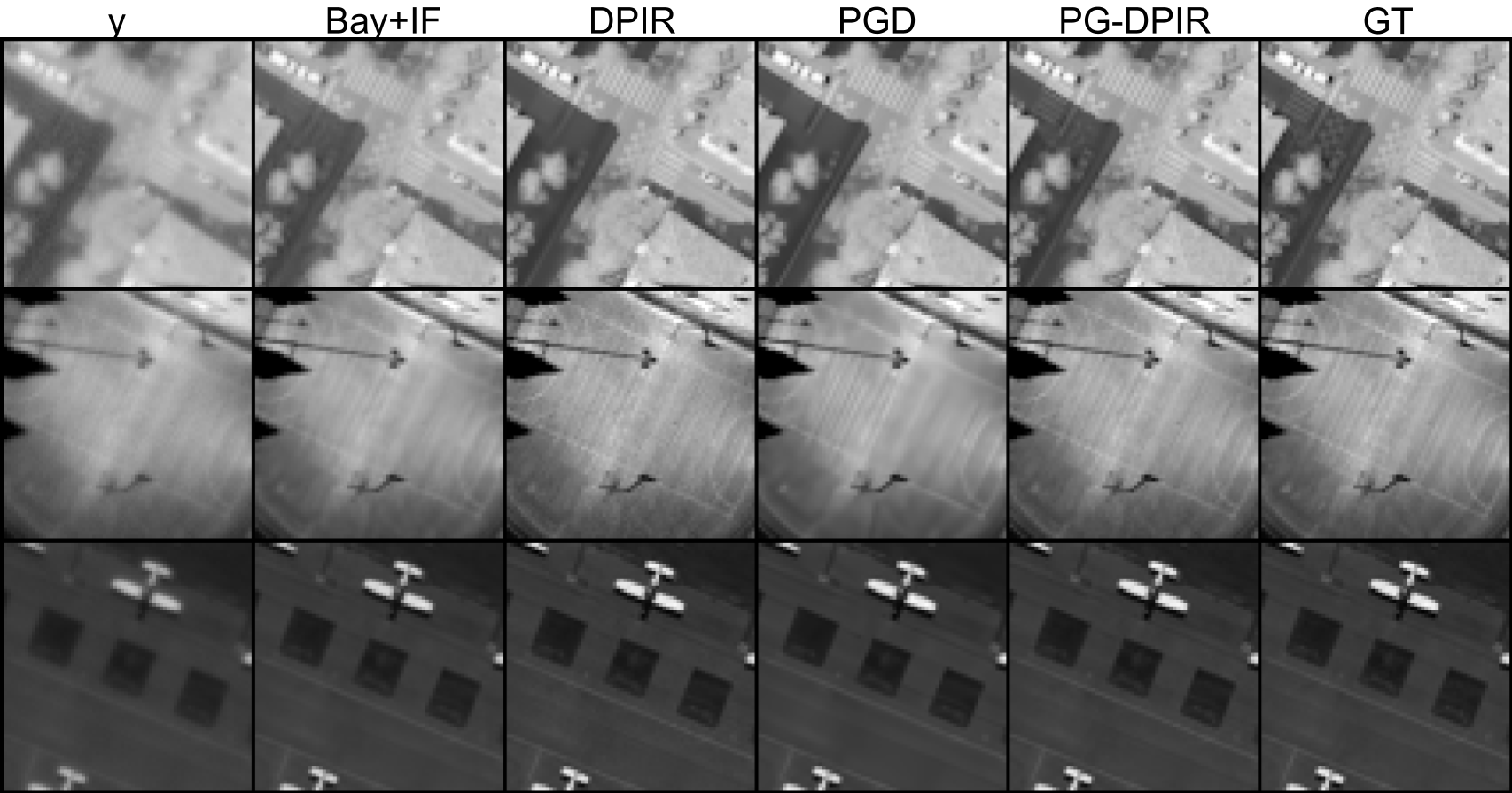}
    \caption{Visual results for the IR problem.}
    \label{fig:50cm}
\end{figure}

The metrics for the two considered inverse problems are provided in \cref{tab:metrics}. First, Bay+IF is outperformed by the deep learning methods. Then, DPIR performs remarkably well given that its forward model is approximate, as it yields better results as PGD. DPIR is also much faster than PGD, illustrating that DPIR is very efficient among the PnP methods.
Last but not least, PG-DPIR outperforms the others methods in terms of metrics, and is almost as fast as DPIR. This shows the interest of the proposed method: the gradient descent to approximate the prox is really fast and does not significantly increase the computational cost of the method. 

\begin{figure}[h]
    \centering
    \includegraphics[width=\linewidth]{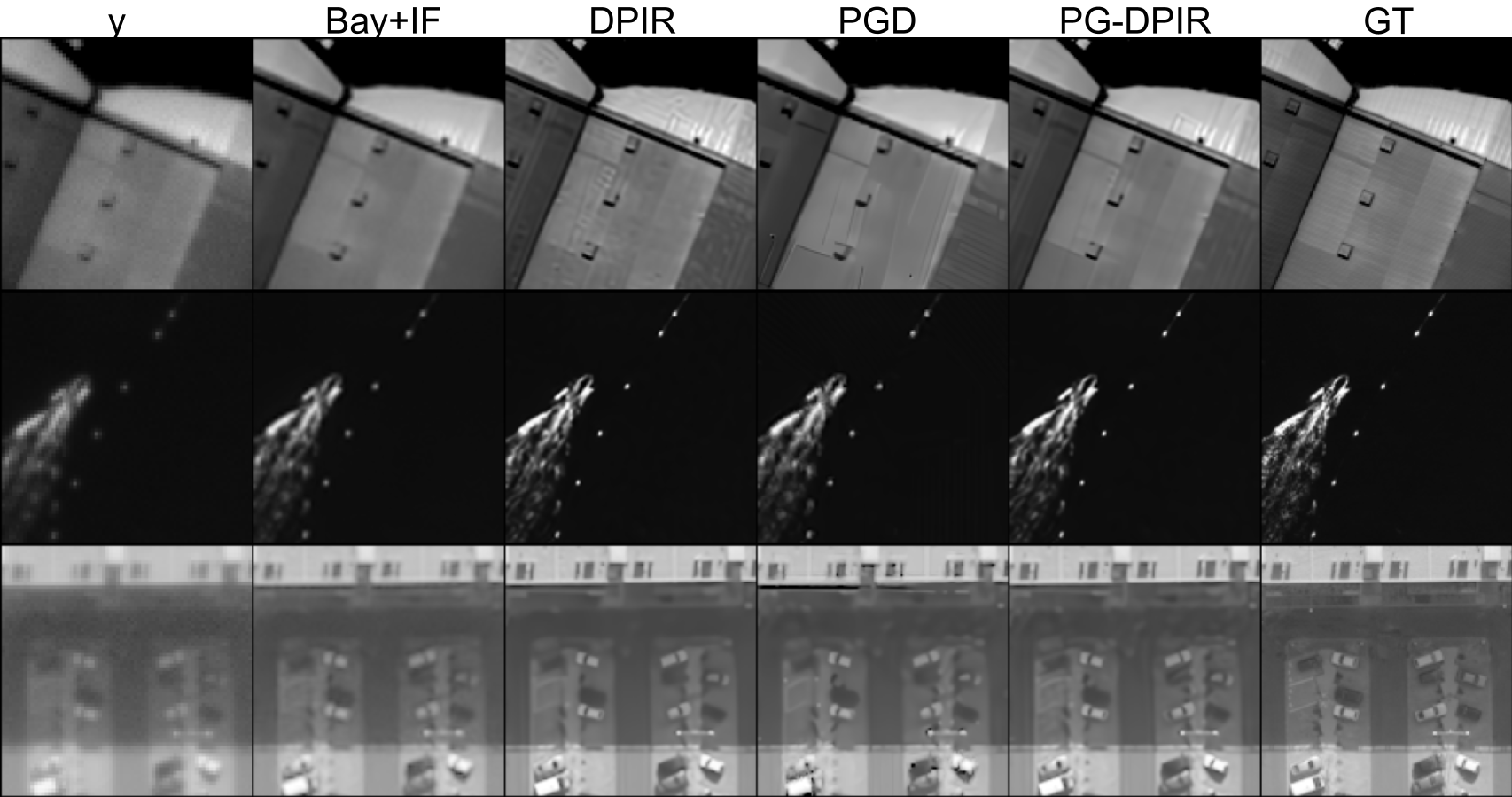}
    \caption{Visual results for the SISR problem.}
    \label{fig:25cm}
\end{figure}

Additionally, visual results are provided for the IR problem in \cref{fig:50cm} and for the SISR problem in \cref{fig:25cm}. The difference between the methods is very clear in the first line of \cref{fig:50cm}: Bay+IF is much smoother as the others, PG-DPIR reconstructs the pedestrian crossing very well, unlike DPIR and PGD. Indeed, PGD seems to regularize too much, while DPIR considers an approximate white Gaussian noise model which seems unable to restore these fine details.  

In \cref{fig:25cm}, PG-DPIR yields less artifacts as PGD and DPIR on the roof, and is more faithful than the other methods in the shadow. These results also highlight that the methods do not really invent information when super resolving by a factor 2, while maintaining a high image quality. Hence, it seems feasible to integrate such a zoom in a processing pipeline to enhance the image before downstream tasks.  

\subsection{Ablation study}

\begin{table}[h]
\caption{Difference between PG-DPIR with the fast proximal initialization (PG-DPIR) or without it (PG-DPIR wout prox\_init). Time denotes the time required to restore one image with a Nvidia Quadro RTX8000 GPU.}
\centering
\begin{tabular}{l|cc|cc}
\hline
 & \multicolumn{2}{c|}{\textbf{IR}} & \multicolumn{2}{c}{\textbf{SISR}} \\ \cline{2-5} 
 & \begin{tabular}[c]{@{}c@{}}PG-DPIR\\ wout prox\_init\end{tabular} & PG-DPIR & \begin{tabular}[c]{@{}c@{}}PG-DPIR\\ wout prox\_init\end{tabular} & PG-DPIR \\ \hline
PSNR $\uparrow$ & 48.70 & 48.66 & 37.19 & 37.18 \\
LPIPS $\downarrow$ & 0.0139 & 0.0138 & 0.1656 & 0.1658 \\
Time & 4.3s & 3.7s & 1090s & 68s \\ \hline
\end{tabular}
\label{tab:ablation}
\end{table}

\Cref{tab:ablation} shows the results of PG-DPIR, with a standard gradient descent to compute the proximal operator (PG-DPIR wout prox\_init), or with the improved gradient descent (PG-DPIR). The two methods have almost the same metrics, but PG-DPIR is faster, especially for the SISR problem. This significant difference makes PG-DPIR usable for real-world satellite image restoration, unlike PG-DPIR wout prox\_init.

\begin{figure}[h]
    \centering
    \includegraphics[width=0.8\linewidth]{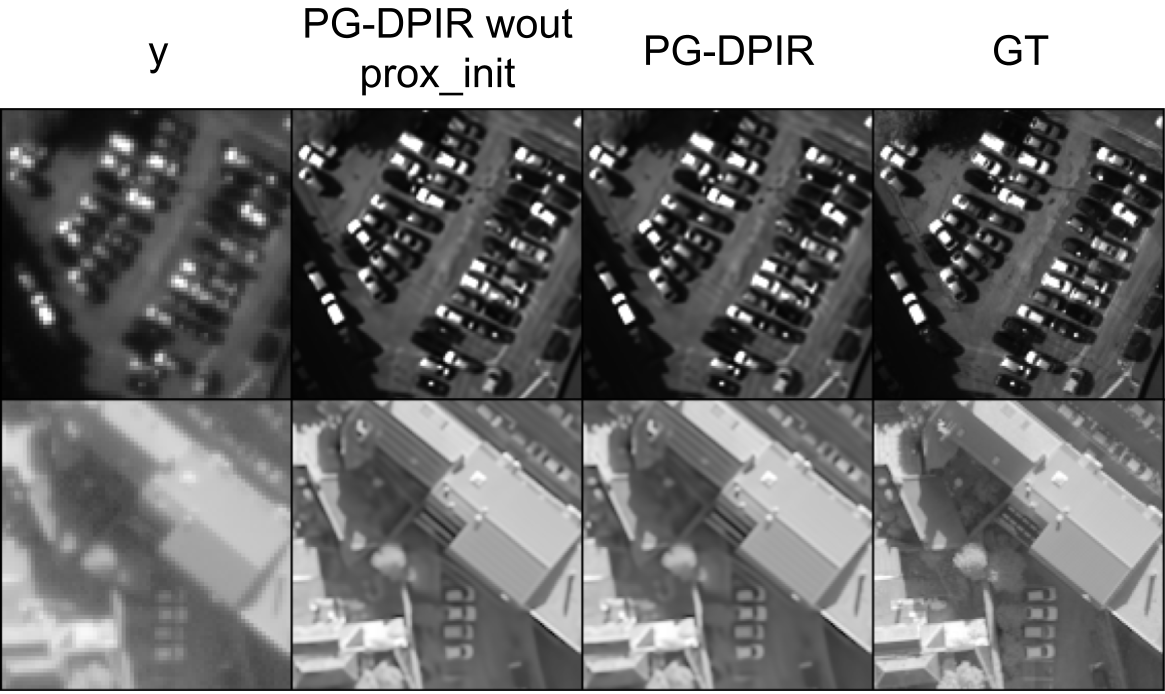}
    \caption{Visual results for the SISR problem with the fast proximal initialization (PG-DPIR) or without it (PG-DPIR wout prox\_init).}
    \label{fig:ablation}
\end{figure}

Visual results for these two variants are given in \cref{fig:ablation}. The two methods do not exhibit visual differences. However, some line artifacts are visible for both methods in the second image. These lines are only visible in SISR and highlight a drawback of our method. It is likely that a more careful choice of the hyperparameters, in the train or in the test process, would erase those artifacts. This is left for future work.   

%% file: sec/3_conclusion.tex
\section{Conclusion}

In this paper, we have introduced PG-DPIR, a plug-and-play method specifically designed for Poisson-Gaussian inverse problems, where the approximate Gaussian noise model remains valid. It employs the Half Quadratic Splitting framework, alternating a data-fidelity proximal step and a denoising step. PG-DPIR improves the speed of the proximal step by initializing the inner gradient descent with an approximate proximal step that can be computed in closed form. This enables to significantly reduce the number of gradient descent iterations required in each proximal step.

We apply PG-DPIR on satellite image restoration problems. We compare PG-DPIR to several baselines on realistic high resolution satellite image restoration problems. These experiments show the interest of our method, as it outperforms the other PnP methods in terms of metrics and computation time. Future work will be dedicated to erasing the remaining artifacts in SISR problems, as well as adapting diffusion-based methods to approximate the Poisson-Gaussian noise model.